\pdfoutput=1

\documentclass[11pt]{article}

\usepackage[preprint]{coling}

\usepackage{times}
\usepackage{latexsym}

\usepackage[T1]{fontenc}

\usepackage[utf8]{inputenc}

\usepackage{microtype}

\usepackage{inconsolata}

\usepackage{graphicx}

\usepackage{multirow}
\usepackage{enumitem,kantlipsum} 

\newcommand\blfootnote[1]{%
  \begingroup
  \renewcommand\thefootnote{}\footnote{#1}%
  \addtocounter{footnote}{-1}%
  \endgroup
}

\title{Adapting Whisper for Regional Dialects: Enhancing Public Services for Vulnerable Populations in the United Kingdom}

\author{Melissa Torgbi\textsuperscript{1}*,
Andrew Clayman\textsuperscript{2}*,
\\
\textbf{Jordan J. Speight\textsuperscript{2}} \and
\textbf{Harish Tayyar Madabushi\textsuperscript{1}}
\\[0.3cm]
\textsuperscript{1} Department of Computer Science, University of Bath, UK \\
\textsuperscript{2} Wyser LTD, UK\\[3mm]
\texttt{\small mat66@bath.ac.uk},
\texttt{\small andrew.clayman@wyser.online} \\[-0.1mm] 
\texttt{\small jordan.speight@wyser.online},
\texttt{\small htm43@bath.ac.uk} \\
\\
}

\begin{document}
\maketitle
\blfootnote{\hspace*{-0.112cm}\textsuperscript{*}Equal Contribution.}
\begin{abstract}

We collect novel data in the public service domain to evaluate the capability of the state-of-the-art automatic speech recognition (ASR) models in capturing regional differences in accents in the United Kingdom (UK), specifically focusing on two accents from Scotland with distinct dialects. This study addresses real-world problems where biased ASR models can lead to miscommunication in public services, disadvantaging individuals with regional accents particularly those in vulnerable populations. We first examine the out-of-the-box performance of the Whisper large-v3 model on a baseline dataset and our data. We then explore the impact of fine-tuning Whisper on the performance in the two UK regions and investigate the effectiveness of existing model evaluation techniques for our real-world application through manual inspection of model errors. We observe that the Whisper model has a higher word error rate (WER) on our test datasets compared to the baseline data and fine-tuning on a given data improves performance on the test dataset with the same domain and accent. The fine-tuned models also appear to show improved performance when applied to the test data outside of the region it was trained on suggesting that fine-tuned models may be transferable within parts of the UK. Our manual analysis of model outputs reveals the benefits and drawbacks of using WER as an evaluation metric and fine-tuning to adapt to regional dialects.

\end{abstract}

\section{Introduction}


Automatic speech recognition (ASR) systems are becoming increasingly embedded in our technologies and processes \cite{koenecke2020racial}. The ease of use of these systems \cite{ibrahim2020study} combined with recent advancements in performance with the use of more sophisticated models makes it particularly appealing for domains with limited resources, including legal areas \cite{trancoso2023impact}, healthcare \cite{latif2020speech} and other public services. As a result, it is important to address potential problems, particularly those that amplify sociolinguistic biases.

Regional and social dialects resulting in speech of the same language having phonological, lexical and grammatical differences present significant challenges for ASR systems \cite{forsberg2003speech}. As English is a high-resource language, there are copious amounts of data available to train ASR models to recognise English. Despite this, many models struggle with variations and dialects of English that are underrepresented in training data \cite{sanabria2023edinburgh}. This phenomenon is observed for multiple variations of English including decreased performance for African American Vernacular English \cite{koenecke2020racial, martin2020understanding}, English as a second language or non-native English \cite{chan2022training, dichristofano2022global} and variations of English within regions including the UK \cite{tatman2017effects, markl2022language}. 

The lack of inclusivity in ASR often leads to disparities between users of these systems \cite{ngueajio2022hey}. As a result, in this work, we investigate the performance of ASR systems on regions in the United Kingdom (UK), specifically areas where accents are less commonly represented in speech datasets. The UK also has socioeconomic links to accent \cite{donnelly2019sociolinguistic, levon2021accent, trudgill1974social}. This is something that may be observed in other countries across languages and so we hope this work will be transferable beyond just English in the UK \cite{bourdieu1991language}.

This research focuses on the state-of-the-art model Whisper \cite{radford2023robust}, a multilingual ASR system that is increasingly used in industry settings. Whisper is trained on a diverse set of 680,000 hours of multilingual data, making it particularly robust for recognising speech across languages, including those with less training data. Whisper is designed to handle real-world audio with noise and challenging conditions better than many existing ASR models. The model demonstrates lower word error rates (WER) compared to earlier models across a variety of benchmarks, including LibriSpeech, Common Voice, and other multilingual datasets \cite{radford2023robust}. Whisper's performance gains have been validated through community usage and industry adoption in particular has motivated the choice to investigate and assess Whisper's capabilities. In this work, we explore Whisper's capabilities to recognise accented speech in public service settings in two areas of the UK, South East Scotland and North East Scotland.

\subsection{Contributions} \label{sec:contributions}
To address the aforementioned challenges, we make the following contributions.

    \begin{enumerate}
        \item[a.] We collect novel data from two real-world public service organisations: a North East Scotland Advice Charity (NESAC) and a South East Scotland Housing Association (SESHA).
        \item[b.] We assess Whisper's performance on the collected data representing two variations of English.
        \item[c.] We fine-tune Whisper to show improved performance on the collected data and the potential transferability of the fine-tuned models to other parts of the UK. 
        \item[d.]  We investigate the evaluation of ASR and the impact of transcription style on the reported performance through manual inspection of model errors highlighting the benefits and drawbacks of using WER as an evaluation metric. 
    \end{enumerate}

We make these contributions with the goal of answering the following research questions.

\begin{enumerate}
    \item How effective is the off-the-shelf state-of-the-art ASR model Whisper in capturing the variations in dialects and accents across regions in the UK?
    \item Is fine-tuning an effective mechanism to adapt models to these dialects?
    \item How good are existing methods of evaluation for real-world applications?
\end{enumerate}

\section{Related Work}

\subsection{Datasets}
Existing research that examines the performance of ASR on variations of English confirms that models struggle with speech that does not match what is most commonly presented
as English in speech corpora \cite{sanabria2023edinburgh, koenecke2020racial, martin2020understanding, chan2022training, dichristofano2022global,tatman2017effects, markl2022language}.
Although some of these studies make their data publicly available, many datasets capture such a broad range of accents that the groups we intend to focus on are not well represented. Our work specifically focuses on accented calls from within the UK. The Open-source Multi-speaker Corpora of the English Accents in the British Isles dataset \cite{demirsahin2020open}, which we use as a baseline dataset in this work, addresses this by collecting data with accents from the British Isles. This dataset, however, does not cover the domains we are interested in and contains scripted speech recorded through a studio microphone rather than spontaneous speech recorded through online calls and phone calls.


\subsection{Fine-tuning}

Fine-tuning is the process of adapting a pre-trained model to new data. Although it has some potential drawbacks including overfitting and catastrophic forgetting, previous work has shown that it is an effective method for improving performance on languages and dialects that are insufficiently represented during pre-training for multiple different models. \citet {zhao2022improving} and \citet{liu2024exploration} show improved performance through fine-tuning for low resource languages using wav2vec \cite{baevski2020wav2vec} and Whisper respectively and \citet{meyer2020artie}
used fine-tuning to improve the performance of DeepSpeech \cite{amodei2016deep} on less common variations of English. We approach variations in English using fine-tuning and investigate how fine-tuned Whisper models perform on two different accents from the UK.  

\section{Data}

We collect new data to assess Whisper's performance on a real-world use case of call transcription. The collected data represents two groups of accents from the UK and consists of calls from two public service scenarios. The real names of these organisations we collect data from have been omitted throughout the paper and replaced with the following representative terminology: North East Scotland Advice Charity (NESAC) and South East Scotland Housing Association (SESHA). These charities provide critical services to the community particularly in vulnerable populations, with a large proportion of callers likely coming from low socio-economic backgrounds vitally in need of these services. NESAC and SESHA offer free legal advice and housing support, respectively, making accurate transcription essential for effective communication and service delivery. Both charities are located in areas with different dialects situated in Scotland. The datasets have been manually annotated with accent labels and manually transcribed for training and comparison with the machine generated transcriptions, we refer to this as the ``human transcript''. We use a subset of our collected data for fine-tuning and the remaining data is reserved for testing. Additionally, we use the Open-source Multi-speaker Corpora of the English Accents in the British Isles dataset \cite{demirsahin2020open} as a baseline dataset for all models.

\subsection{Data Privacy and Ethics} 
Given the sensitive nature of the data involved, we take extra care to ensure its handling is secure and ethically sound (Also see Section \ref{sec:ethics}). This research was conducted in collaboration with a licensed transcription service provider for the aforementioned public service organisations. All data collection adhered strictly to local and regional legal and regulatory requirements. The data is used specifically to reduce potential biases in the services provided to these organisations, ensuring its appropriate and justified use. Collected data is securely stored on encrypted servers and is destroyed within a three-month period, as mandated by the relevant regulations. All personnel who have access to private data are bound by agreements to safeguard data privacy. Personnel who do not require access to private data worked with publicly available datasets, and insights from their analyses are shared with authorised personnel for implementation. These measures ensured that private data remained secure and is used solely to reduce biases in the transcription services provided. 

\subsection{North East Scotland Advice Charity}
The North East Scotland Advice Charity data, or NESAC, contains calls between community members and advisors. These calls span numerous topics including debt and financial advice, welfare benefits, housing and tenancy issues, employment issues, consumer rights, legal advice, relationship issues, immigration and residency. Transcripts generated from these calls will then be used by the organisation for downstream tasks including the creation of a transcript summary for documentation and client follow-up. Given that the content of the call contains critical information, it is essential that the transcription is accurate as errors or omissions could negatively affect the caller's well-being. Tables \ref{tab:NESACaccent} and \ref{tab:NESACgender} show the split of the collected NESAC data by accent and gender.

\begin{table}[ht!]
\centering
\begin{tabular}{llll}
\hline
\textbf{Accent} & \textbf{Advisors} & \textbf{Callers}\\
\hline
Scottish & 93.75 & 78.13\\
English & 3.13 & 12.50\\
Other & 3.13 & 9.38\\
\end{tabular}%
\caption{Percentage of accents in the NESAC dataset.}
\label{tab:NESACaccent}
\end{table}

\begin{table}[ht!]
\centering
\begin{tabular}{llll}
\hline
\textbf{Speaker} & \textbf{Female} & \textbf{Male} & \textbf{Unknown}\\
\hline
Caller & 43.75 & 56.25 & 0.00  \\
Advisor & 71.88 & 25.00 & 3.13 \\
\end{tabular}%
\caption{Percentage of genders in the NESAC dataset.}
\label{tab:NESACgender}
\end{table}

\subsection{South East Scotland Housing Association}
The South East Scotland Housing Association data, or SESHA, contains calls with advisors related to housing and properties provided by the South East Scotland Housing Association charity. The calls typically include conversations about whether someone is eligible to obtain a home through them, if they can join the waiting list for a home, change home, or file a complaint about a neighbour. Similar to NESAC, these calls are transcribed and used by the organisation for other tasks such as summarising the transcripts for documentation and client follow-up. The vitality of accurate transcription also applies here due to the risk of error or missing information resulting in well-being concerns for the caller. Tables \ref{tab:ehlaaccent} and \ref{tab:SESHAgender} show this data split by accent and gender.

\begin{table}[htbp!]
\centering
\begin{tabular}{llll}
\hline
\textbf{Accent} & \textbf{Advisors} & \textbf{Callers}\\
\hline
Scottish & 80.69 & 92.84\\
English & 18.42 & 2.37 \\
Irish & 0.87 & 0.65\\
Other & 0.00 & 3.90\\
\end{tabular}%
\caption{Percentage of accents in the SESHA dataset.}
\label{tab:ehlaaccent}
\end{table}


\begin{table}[htbp!]
\centering
\begin{tabular}{lll}
\hline
\textbf{Speaker} & \textbf{Female} & \textbf{Male}\\
\hline
Caller & 72.51 & 27.49 \\
Advisor & 81.78 & 18.22  \\
\end{tabular}%
\caption{Percentage of genders in the SESHA dataset.}
\label{tab:SESHAgender}
\end{table}


\section{Experimental Setup}\label{sec:setup}

To address the research questions outlined in Section \ref{sec:contributions}. We run two experiments and a manual analysis. The first experiment looks at the effectiveness of Whisper in capturing variations in dialect in the UK and the second explores fine-tuning as a mechanism to adapt the Whisper model to accents. Finally, we conduct a manual analysis of model errors to better understand the effectiveness of our chosen evaluation metric WER. This section describes the experimental setup for these experiments.

We test the Whisper large-v3 model on a subset of our NESAC and SESHA datasets where each test set has approximately 5 hours of data. The large-v3 model for Whisper was selected over the other sizes available as it gave the best performance in our initial experiments. 

Whisper large-v3 is also used as a base model in our fine-tuning experiment. We fine-tune two models, one using NESAC and the other using SESHA. The same two test sets from the first experiment are used to evaluate the performance of the fine-tuned models as the training and test data were separated before fine-tuning. For the training of the fine-tuned models a learning rate of 5x$10^{-6}$ and a batchsize of 64 were used with 47 hours of the NESAC data used to train the NESAC fine-tuned model and 46 hours of the SESHA data to train the SESHA fine-tuned model. 

\section{Experiment 1: Whisper}

\begin{figure}[ht!]
  \includegraphics[width=\columnwidth]{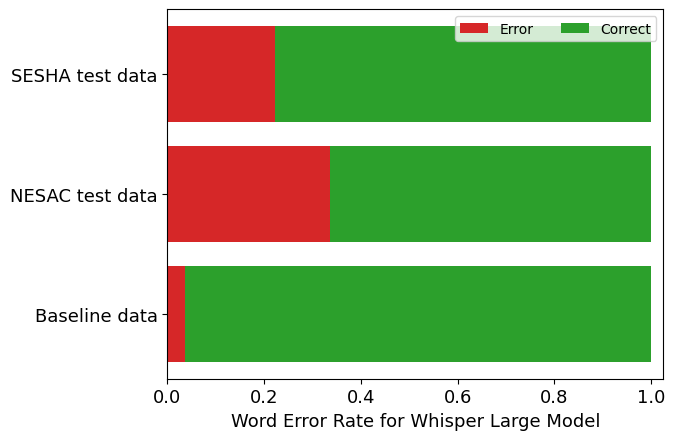}
  \caption{Word error rate of the Whisper large-v3 model on the baseline dataset and two test datasets NESAC test data and SESHA test data.}
  \label{fig:whisperwer}
\end{figure}

To answer Research Question 1 outlined in Section \ref{sec:contributions}, this experiment focuses on the out-of-the-box performance of the Whisper large-v3 model on our collected data representing accents from North East Scotland captured in NESAC and South East Scotland captured in SESHA. The results of this experiment are shown in Figure \ref{fig:whisperwer} and the first row of Table \ref{tab:results}.

\subsection{Empirical Evaluation and Analysis}

The performance of the Whisper large model on the baseline dataset and test dataset is shown in figure \ref{fig:whisperwer}. Whisper performs well on the baseline data achieving a WER of 3.64\% whereas it does comparatively worse on our test datasets, NESAC and SESHA. This is a difference that is observed for the fine-tuned models in Experiment 2 as well although not to the same extent as the Whisper large model. Since the baseline data is open source, there is a possibility that this data may have featured in the pre-training data for Whisper. The difference in performance could also suggest that our data is more difficult to transcribe than the baseline data. This may be due to a number of factors including accent, dialect, domain-specific language, quality of the calls, and the conversational nature of the calls in the test data compared to the baseline data that involves participants to read aloud. Some of the difference in performance may also be due to transcription style. This is something we explore further in Section \ref{sec:manualanalysis}.

\section{Experiment 2: Fine-tuned Models}

\begin{figure}[ht!]
  \includegraphics[width=\columnwidth]{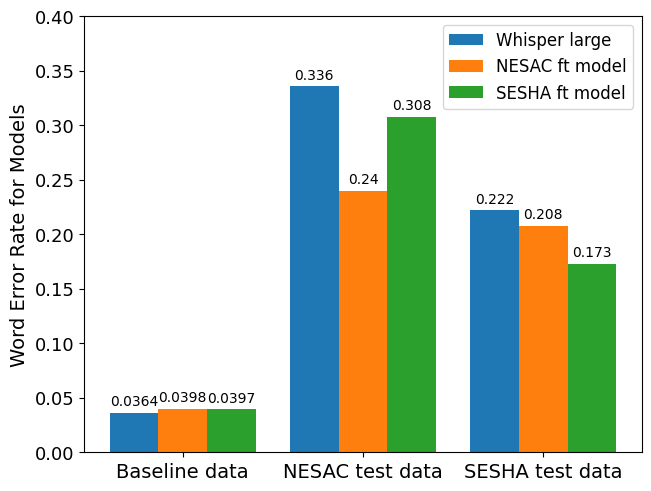}
  \caption{WER of the Whisper large-v3 model, the NESAC fine-tuned model and SESHA fine-tuned model on the baseline dataset and two test datasets NESAC test data and SESHA test data.}
  \label{fig:datawer}
\end{figure}

To answer Research Question 2 outlined in Section \ref{sec:contributions}, this experiment investigates the effectiveness of fine-tuning for improving the performance of Whisper on our accented public service test datasets NESAC and SESHA. We fine-tune two models using the settings described in Section \ref{sec:setup}. Figure \ref{fig:datawer} and Table \ref{tab:results} compare the performance of the Whisper large model and the two fine-tuned models where "NESAC ft model" is fine-tuned on our NESAC training data and "SESHA ft model" is fine-tuned on the SESHA training dataset.

\begin{table}[ht!]
\centering
\begin{tabular}{lp{1.2cm}p{1cm}p{1cm}}
\hline
\textbf{Model} & \textbf{Baseline data} & \textbf{NESAC test data} & \textbf{SESHA test data} \\
\hline
Whisper large &	0.0364 &	0.336 &	0.222 \\
NESAC ft model &	0.0398 &	0.240 &	0.208 \\
SESHA ft model & 	0.0397 &	0.308 &	0.173 \\

\end{tabular}
\caption{WER of the Whisper large-v3 model and the fine-tuned NESAC and SESHA models on the baseline dataset and two test datasets NESAC and SESHA.}
\label{tab:results}
\end{table}

\subsection{Empirical Evaluation and Analysis}

The results of this experiment comparing the performance of the models on the baseline dataset show that although the Whisper model has the lowest WER, all three models have comparable performance on the baseline data. 

Looking at performance on our accented test data, the models that perform the best on each test set are the models that are fine-tuned on the data that matches the test. For the NESAC test data, the NESAC fine-tuned model performs the best, followed by the SESHA fine-tuned model and then the Whisper large model. Similarly, for the SESHA test data, the SESHA fine-tuned model performs the best, followed by the NESAC fine-tuned model and then the Whisper large model. This suggests that although NESAC and SESHA contain distinct dialects, the models may be picking up on similarities in dialect resulting in better performance than the Whisper large model. The Whisper large model performs the worst for each test data set. This may be due to less familiar dialects or domain-specific language. We explore this further by conducting a manual analysis of each model’s errors.

\section{Manual Analysis} \label{sec:manualanalysis}
To address Research Question 3 outlined in Section \ref{sec:contributions} and better understand the effectiveness of WER as an evaluation metric for our models, we manually inspect a portion of the errors from each model on the baseline data as well as the NESAC and SESHA test data. Since the NESAC and SESHA datasets contain sensitive information, we mostly present our findings with examples from the baseline dataset. Although the fine-tuned models exhibited higher WER on the baseline data compared to the Whisper large model, our manual analysis suggests that this does not necessarily indicate a worse performance.

\subsection{Baseline Data Error Analysis}

After manually inspecting randomly selected errors from each model, we found a few common transcription style differences that were picked up as errors. These errors include having spaces in different places, spelling variations of words, mistakes that are corrected in speech (reparandum) and differences in ways of recording time. These errors along with examples from the baseline dataset are presented in Table \ref{tab:errorsset1}.

\begin{table*}[htp!]
\centering
\begin{tabular}{p{2cm}p{2.7cm}p{9.3cm}}
\hline
\textbf{Error Type} & \textbf{Transcript} & \textbf{Content} \\
\hline
\multirow{3}{=}{Spacing } & Human & take the \textbf{south eastern} main line from charing cross station \\
& Whisper & take the \textbf{south eastern} main line from charing cross station \\
& NESAC ft model & take the \textbf{southeastern} main line from charing cross station \\
\hline
\multirow{4}{=}{Common noun homophone }  & Human & the participating officers exchanged flasks of \textbf{whisky} and vodka \\
& Whisper & the participating officers exchanged flasks of \textbf{whisky} and vodka \\
& NESAC ft model & the participating officers exchange flasks of \textbf{whiskey} and vodka \\
& SESHA ft model& the participating officers exchange flasks of \textbf{whiskey} and vodka \\
\hline
\multirow{8}{=}{ Reparandum} & Human & concentrated solar power uses molten salt energy storage in a tower or in trough configurations\\
& Whisper & concentrated solar power uses molten salt energy storage in a tower or in trough configurations\\
& NESAC ft model & concentrated solar power uses molten salt energy storage in a tower or in trough \textbf{sorry trough} configurations \\
& SESHA ft model & concentrated solar power uses molten salt energy storage in a tower or in trough \textbf{sorry trough} configurations\\
\hline
\multirow{7}{=}{Date/Time Formatting}
&Human & before that on april \textbf{the} 7th at \textbf{half past 10} you had rob is birthday gathering\\
& Whisper & before that on april \textbf{the} 7th at \textbf{half past 10} you had rob is birthday gathering\\
& NESAC ft model & before that on april 7th at \textbf{10.30} you had rob is birthday gathering\\
& SESHA ft model & before that on april 7th at \textbf{10.30 pm} you had rob is birthday gathering\\
\end{tabular}
\caption{Examples where the fine-tuned model gets it wrong, and the Whisper large model gets it right, but the errors are trivial, where it does not affect the content of the text or even a human may get it wrong.}
\label{tab:errorsset1}
\end{table*}

\begin{table*}[htp!]
\centering
\begin{tabular}{p{2cm}p{2.7cm}p{9.3cm}}
\hline
\textbf{Error Type} & \textbf{Transcript} & \textbf{Content} \\
\hline
\multirow{6}{=}{Contextual Bias} & Human & \textbf{mutually} assured destruction is a doctrine of military strategy and national security policy\\
& Whisper & \textbf{mutually} assured destruction is a doctrine of military strategy and national security policy\\
& SESHA ft model & \textbf{neutrally} assured destruction is a doctrine of military strategy and national security policy\\
\hline
\multirow{3}{=}{Contextual Bias}& Human & making a phone call to \textbf{courtney}\\
&Whisper & making a phone call to \textbf{courtney}\\
& NESAC ft model & making a phone call to \textbf{court name}\\
\hline
\multirow{4}{=}{Contextual Bias}& Human & yes it is \textbf{snowing} in copenhagen\\
& Whisper & yes it is \textbf{snowing} in copenhagen\\
& NESAC ft model & yes it is \textbf{now ending} in copenhagen\\
& SESHA ft model & yes it is \textbf{9} in copenhagen\\
\end{tabular}
\caption{Evidence of a loss of contextualisation or real mistakes, where the fine-tuned model is wrong and the Whisper large model is right.}
\label{tab:errors_introduced_by_fine-tuning}
\end{table*}

\begin{table*}[ht!]
\centering
\begin{tabular}{p{2cm}p{2.7cm}p{9.3cm}}
\hline
\textbf{Error Type} & \textbf{Transcript} & \textbf{Content} \\
\hline
\multirow{3}{=}{Phonetic discrimination} & Human & a \textbf{bored} cat laying on a couch\\
& Whisper & a \textbf{bald} cat laying on a couch\\
& NESAC ft model & a \textbf{bored} cat laying on a couch\\
& SESHA ft model & a \textbf{bored} cat laying on a couch\\
\hline
\multirow{4}{=}{Proper noun} & Human & it is 18 degrees with a chance of showers in \textbf{cambuslang}\\
& Whisper & it is 18 degrees with a chance of showers and \textbf{canvas lying}\\ 
&NESAC ft model &it is 18 degrees with a chance of showers in \textbf{cambuslang}
\end{tabular}
\caption{Examples where the Whisper large model gets it wrong, and the fine-tuned models get it right showing evidence of tuning to UK accents or understanding place names.} \label{tab:whispererrors}
\end{table*}

We also identified cases where the fine-tuned models made errors that the Whisper model did not, and vice versa. These additional examples are shown in Tables \ref{tab:errors_introduced_by_fine-tuning} and \ref{tab:whispererrors}.

We applied several post-processing steps to the baseline data transcripts that address some of the common errors caused by differences in transcription style to observe the impact on WER. Spacing errors were initially addressed by adding a space at every possible position in an utterance, keeping the change only if it reduced the WER. An alternative approach involved removing spaces between words where it increased the alignment between the human transcript and the ASR model’s output. Additionally, we found that although the baseline dataset contains accents from the UK, the human transcript contained American spellings of words whereas our model training data contains British spellings. Addressing the American spellings involved replacing occurrences of "ize" and "zation" with "ise" and "sation". Adjustments to dates were also made using regular expressions to capture dates in the format “the 5th of January” and converted them to “5th January” to match the transcription style. By normalising these transcription style differences, we aimed to create a fairer comparison between the models.

\begin{figure}[ht!]
  \includegraphics[width=\columnwidth]{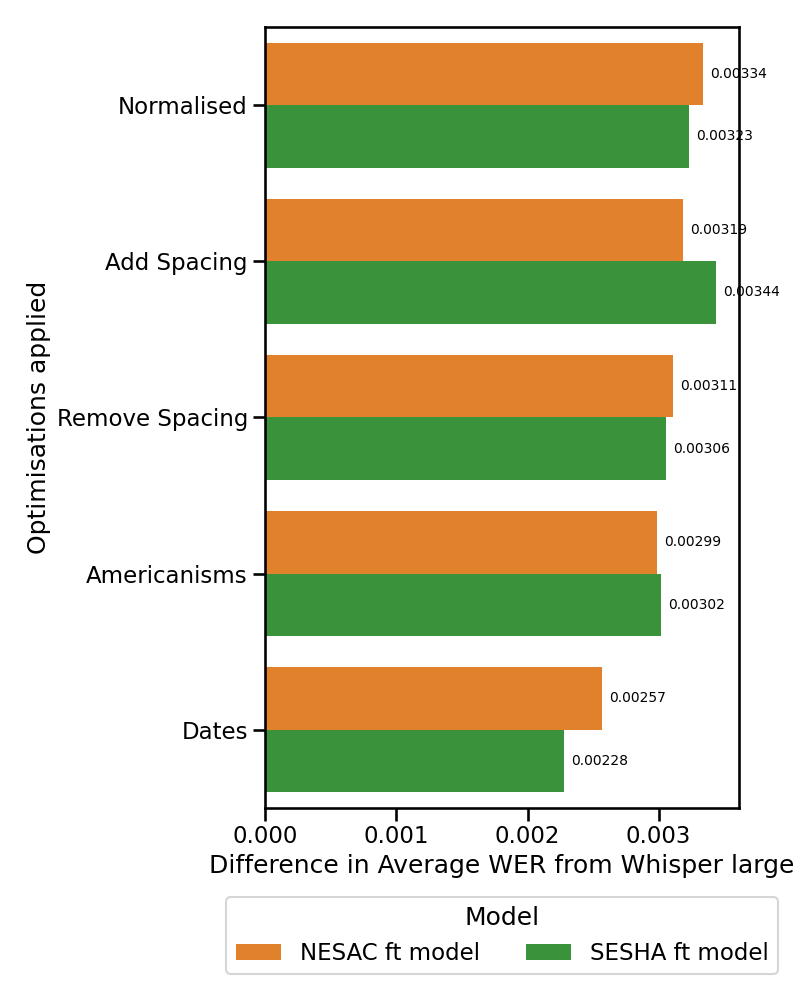}
  \caption{Difference in average word error rate (WER) from Whisper large-v3 after cumulative automated optimisation steps.}
  \label{fig:optimised_WER}
\end{figure}

Figure \ref{fig:optimised_WER} shows a graph that illustrates the automated normalisation steps applied to address a higher WER due to spacing errors, date formats, and American spellings in the human transcripts. Applying these post-processing optimisation steps also improved the WER for the Whisper large model, however, we are particularly interested in the difference in the performance of the fine-tuned models compared with Whisper large. Consequently, Figure \ref{fig:optimised_WER} shows the difference in average WER of the NESAC and SESHA fine-tuned models when compared to the Whisper large model with the same post-processing applied to the human transcript.  The post-processing optimisations are cumulative, so the lower bars have had all the previous optimisations applied. We observe that the cumulative effect of all the post-processing optimisations closes the gap in performance between the Whisper model and our fine-tuned models on the baseline dataset.

This suggests that the higher WER observed initially was largely due to transcription style discrepancies rather than actual recognition errors.

These findings indicate that the fine-tuned models are indeed improving in their ability to understand the target accents and proper nouns, even if this improvement is not fully captured by WER due to transcription style differences and occasional errors.

We also identified cases where the fine-tuned models made errors not present in the Whisper model. These errors are shown in Table \ref{tab:errors_introduced_by_fine-tuning}.

From these examples, it appears that the fine-tuning process may have introduced some contextual bias, leading to a loss of contextual understanding in everyday speech. For instance, in the first example, the SESHA fine-tuned model transcribed "neutrally assured destruction" instead of the correct "mutually assured destruction". The Whisper large model correctly transcribed "mutually", likely due to its broader contextual understanding of common phrases in military strategy.

This suggests that while the fine-tuned models are improving in recognising accent-specific vocabulary and slang, such as 'aye' or 'dinnae,' they may become overly sensitive to certain phonetic patterns at the expense of general language comprehension. The fine-tuning might have made the models more verbatim in transcribing accent-specific pronunciations, causing them to misinterpret words that require contextual cues for accurate transcription.

Similarly, in the second example, the NESAC fine-tuned model misrecognised "courtney" as "court name," and in the third example, both the NESAC and SESHA fine-tuned models misheard "snowing" as "now ending" and "9," respectively. These errors indicate potential overfitting to the accent-specific data, where the models prioritise phonetic patterns common in the fine-tuning datasets over contextual understanding. 

These findings imply that the fine-tuned models may exhibit a trade-off between improved accent comprehension and maintaining contextual accuracy in everyday speech. The introduction of contextual bias through fine-tuning highlights the need for a balanced approach that enhances accent recognition without compromising the models' ability to utilize context for accurate transcription.

Overall, our manual analysis suggests that while the fine-tuned models may show a higher WER, this metric does not fully reflect their enhanced performance in accent comprehension and transcription accuracy for certain types of content. However, it also reveals areas where fine-tuning may inadvertently reduce the models' contextual understanding, indicating a need for careful balancing during the fine-tuning process.

\subsection{Test Data Error Analysis}

In evaluating the performance of our fine-tuned Whisper models on the NESAC and SESHA test datasets, we observed that both fine-tuned models outperformed the Whisper large model across both datasets. Notably, the fine-tuned models achieved the highest performance on the dataset they were specifically trained on, highlighting the effectiveness of the fine-tuning process in adapting to the unique characteristics of the target data.

However, a significant portion of the errors identified during manual analysis were attributable to transcription style differences rather than genuine recognition inaccuracies. For instance, variations such as "all right" versus "alright" were frequently noted, where the models correctly transcribed the spoken words but differed in transcription conventions. These discrepancies do not indicate a decline in the models' recognition capabilities but rather reflect differences in transcription preferences or standards.

Additionally, other transcription style variations, such as the use of regional colloquialisms, handling of filler words like "um" or "uh," and differences in formatting dates and times, contributed to the error counts. These factors can artificially inflate the WER without representing actual misrecognitions, underscoring the limitations of relying solely on WER as an evaluation metric.

Despite these transcription style discrepancies, the fine-tuned models demonstrated enhanced understanding of accent-specific pronunciations and regional vocabulary. For example, in instances where the Whisper large model misrecognised words due to accent variations, the fine-tuned models accurately captured the intended words. Some examples of this are the Whisper large model transcribing 'moment' when the word is 'minute', 'that'll' when it should be 'I'll', 'email' instead of 'female', as well as other similar mistakes. This improvement suggests that the fine-tuning process not only aligns the models with the transcription style of the training data but also enhances their ability to comprehend and accurately transcribe speech with specific accent characteristics.

Furthermore, the fine-tuned models were better at managing colloquial expressions and regional terminology present in the NESAC and SESHA test datasets. This indicates that while WER is a useful quantitative metric, it does not fully account for the models' improved capabilities in understanding accented speech and adapting to varied transcription styles.

Overall, our manual error analysis reveals that the fine-tuned Whisper models offer superior performance in accurately transcribing speech from the NESAC and SESHA test datasets. The higher WER observed is largely a result of transcription style differences rather than a decline in recognition quality. This underscores the importance of supplementing quantitative metrics like WER with qualitative analyses to gain a comprehensive understanding of ASR model performance, especially in diverse and real-world settings.



\section{Conclusion and Future Work}

 

This work uses a novel dataset to assess Whisper's ability to recognise speech from two dialects in the UK. We evaluate Whisper large and fine-tuned versions of the model on a baseline dataset and our two test datasets. We find that all of the models have worse performance on our North East Scottish and South East Scottish test data compared to the baseline data, the Whisper model performs better when it is fine-tuned and tested on data from the same distribution and there may be evidence of dialect transferability for our fine-tuned models. We conducted a manual analysis of the errors from each model and found that differences in transcription style appear to negatively impact the observed WER. The manual analysis also demonstrated evidence of the fine-tuned models successfully adapting to the target dialect as well as cases where the fine-tuning approach negatively impacted the models' contextual understanding. This indicates the need for a careful balance during the fine-tuning process and highlights both the potential and the drawback of using fine-tuning for variations in English in public services for vulnerable populations.

We hope to investigate the transferability of fine-tuned Whisper models further in future work by collecting more data that represents a wider range of accents from within the UK and evaluate the transferability of fine-tuned models on accents from these other regions. Furthermore, we aim to incorporate approaches that avoid the use of confidential and sensitive data, which NESAC and SESHA are in this case.

\section{Limitations}
In this research, we collect novel data to investigate the ability of fine-tuning and Whisper large to adapt to accents in the UK in a real-world public service setting. Despite our best efforts annotation bias may persist in our work, this however further emphasises the need for manual analysis in our approach. In this research, we only look at two accents but it would be advantageous if we were able to collect more data that had a broader range of UK accents represented in the two public service areas we explore. We only explore fine-tuning as a method to address variations in English but we choose this method over others for generalisability as fine-tuning is a technique that can be applied to other pre-trained models. We also only intentionally look at English. Although we believe this work may be applicable to multiple languages this is something that should be tested across other languages. The sensitive nature of our collected data has also meant that we are unable to publicly share the data. Nonetheless, this work highlights both the potential and the drawback of using Whisper, fine-tuning and WER for variations in English. 

\section{Ethics}
\label{sec:ethics}
This work was done in collaboration with government sanctioned organisations that provide legal and housing support within the UK. These are established structures that we cannot name for legal reasons. Their recording of calls is strictly governed by GDPR and other legal frameworks and goes through an independent audit process. We collect data from them after careful legal and ethical reviews. This research was funded by the EPSRC and therefore underwent additional scrutiny with strict legal and ethical framework to ensure the security and privacy of these calls, and is also audited. The sections relevant to the analysis also underwent ethical review at the university partner. People working on this industry led project are trained to work with private information. This information remains on the company’s servers at all times and the research institute only works on publicly available data, transferring research methods and ideas to the industry led partner to ensure privacy. All transcripts are permanently deleted after a fixed time period. The datasets were manually transcribed. We hired UK-based professional annotators who follow professional standards to transcribe the audio and label accents.

\section*{Acknowledgments}
This work was supported by Innovate UK [grant number 10093501] through a Collaborative R\&D grant.

\bibliography{custom}

\appendix

\end{document}